# A LOGICAL INTERPRETATION OF DEMPSTER SHAFER THEORY, WITH APPLICATION TO VISUAL RECOGNITION

Gregory M. Provan*
Department of Computer Science
University of British Columbia
Vancouver, BC
Canada V6T 1W5

## Abstract

We formulate Dempster Shafer Belief functions in terms of Propositional Logic, using the implicit notion of provability underlying Dempster Shafer Theory. The assignment of weights to the propositional literals enables the Belief functions to be explicitly computed using Network Reliability techniques. Also, the updating of Belief functions using Dempster's Rule of Combination corresponds to incremental updating of the corresponding support clauses. This analysis formalizes the implementation of Belief functions within an ATMS. We describe VICTORS, a visual recognition system based on an ATMS extended with Belief functions. Without Dempster Shafer theory, VICTORS computes all possible visual interpretations (i.e. all logical models) without discriminating the best interpretations. Incorporating Dempster Shafer theory enables optimal visual interpretations to be computed and a logical semantics to be maintained.

## 1 INTRODUCTION

Dempster Shafer (DS) Theory has been proposed as a calculus for reasoning under uncertainty to rival Probability Theory in expressive power and effectiveness. The DS Belief function is based on the notion of the provability of a proposition $\theta$ in terms of its subsets [14]. In this paper we explicitly define DS Theory in terms of Propositional Logic, using this implicit notion of provability underlying DS Theory. Hence, we describe both how Dempster Shafer Theory can be assigned a logical semantics and propositional logic can be extended with an uncertainty calculus.

We assume an assignment $\varrho : x \to [0,1]$ to a set $x$ of literals which corresponds to the set of focal propositions, and define a set of clauses $X = \{X_1, ..., X_m\}$ which denote the provability relations underlying the power set $\mathcal{P}\Theta$ of a set $\Theta$ of propositions. We show that the support set for a clause $X_i$ with respect to the set $X$ of propositional clauses, $\xi(X_i, X)$, corresponds to a symbolic representation of the DS Belief function for $X_i$. Explicitly computing the numerical value for the Boolean formula for $Bel(X_i)$, $\xi(X_i, X)$, is equivalent to the evaluation of the network reliability of a network defined by $\xi(X_i, X)$. Moreover, we show that the pooling of information, which in DS Theory is represented as $Bel(\theta) = Bel' \oplus Bel''$, corresponds in our logical formulation to support set updating.

In addition to exploring the underlying relationship between DS Theory and propositional logic, we also briefly examine an implementation of DS Theory based on this logical formulation using an Assumption-based Truth Maintenance System (ATMS) [3]. We describe the application of an ATMS extended with DS theory to a model-based visual recognition problem, as implemented within a system called VICTORS.

The remainder of the paper is organized as follows. Section 2 briefly defines several important concepts in DS Theory. Section 3 introduces the logical notation. Then in Section 4 we define DS Theory in terms of this notation. Section 5 examines an implementation of DS Theory based on ATMS. We use a visual recognition problem as an example of the application of this ATMS-based implementation, as discussed in Section 6. Finally, in Section 7 we state our conclusions.

## 2 DEMPSTER SHAFER THEORY REVIEW

Many good descriptions of Dempster-Shafer (DS) theory exist, e.g. [4], [14]. We assume familiarity with DS theory, state a few basic relationships, and refer the reader to the references.

In DS theory, weights are assigned to subsets as well as elements of a mutually exclusive set of focal propositions $\Theta$. A mass function $\varrho : 2^\Theta \to [0,1]$ assigns weights to subsets $\theta$ of $\Theta$ subject to the following properties: $\varrho(\theta) \in [0,1]$, $\sum_{\theta \subseteq \Theta} \varrho(\theta) = 1$ and $\varrho(\emptyset) = 0$.

One measure in DS theory which is derived from this mass function is *Belief*, the degree of belief in proposition subsets from which a proposition $\theta$ can be proven:

$$Bel(\theta) = \sum_{\varphi \subseteq \theta} \varrho(\varphi). \qquad (1)$$

*The author completed this research with the support of the University of British Columbia Center for Integrated Computer Systems Research, BC Advanced Systems Institute and NSERC grants to A.K. Mackworth.

Dempster's Rule of Combination defines an updated mass function for a proposition $\theta$ provable in terms of $\theta_i$ and $\theta_j$, for all $\theta, \theta_i, \theta_j \subseteq \Theta$, as:

$$\varrho'(\theta) = \frac{\sum_{i,j: \theta_i \cap \theta_j = \theta} \varrho_1(\theta_i) \varrho_2(\theta_j)}{1 - \sum_{i,j: \theta_i \cap \theta_j = \emptyset} \varrho_1(\theta_i) \varrho_2(\theta_j)}. \quad (2)$$

The Belief function $\varrho'$ is also denoted as $Bel_1 \oplus Bel_2$. The numerator assumes independence of propositions. Viewed in set-theoretic terms, this is simply "summing" the mass functions of all sets in which $\theta$ is provable. The denominator of Equation 2 is a normalizing term, given that DS Belief is assigned only to non-contradictory subsets.

## 3 PROPOSITIONAL LOGIC REVIEW

We use a propositional language which contains a finite set of propositional symbols and the connectives $\vee, \wedge, \neg,$ and $\Rightarrow$. A propositional *literal* is a propositional symbol or its negation. $\mathbf{x} = \{x_1, \overline{x_1}, ..., x_n\}$ is a set of propositional literals. A *clause* is a finite disjunction of propositional literals, with no repeated literals. $X = \{X_1, ..., X_l\}$ is a set of input clauses. [1]

A *Horn clause* is a clause with at most one unnegated literal. For example, a Horn-clause $X_i$ can be written as $\overline{x_1} \vee \overline{x_2} \vee \overline{x_3} \vee ... \vee \overline{x_k} \vee x, k \geq 0$.

A *prime implicate* of a set $X$ of clauses is a clause $\pi$ (often called $\pi(X)$ to denote the set $X$ of clauses for which this is a prime implicate) such that (1) $X \models \pi$, and (2) for no proper subset $\pi'$ of $\pi$ does $X \models \pi'$. We denote the set of prime implicates with respect to $X$ by $\Pi(X)$.

$\xi_j$ is the $j^{th}$ *support clause* for $x$ with respect to $X$ (often called $\xi_j(x, X)$) iff (1) $X \not\models \xi_j$, (2) $x \cup \xi_j$ does not contain a complementary pair of literals (i.e. both $x_j$ and $\overline{x_j}$), and (3) $X \models x \cup \xi_j$. The *set of support* for a literal $x$ is the disjunction of the support clauses for $x$, i.e. $\xi(x, X) = \bigvee_i \xi_i(x, X)$.

We call the conjunction of the $X_i$'s a Boolean expression $F$, i.e. $F = \bigwedge_{i=1,...,l} X_i$.

## 4 DEMPSTER SHAFER THEORY FORMULATION IN LOGIC-BASED TERMS

Shafer [14] implicitly defined a correspondence between set-theoretic notions relevant to subsets of $\Theta$ and logical notions. More precisely, as described on page 37 of [14], if $\theta_1$ and $\theta_2$ are two subsets of $\Theta$ and $x_1$ and $x_2$ are the corresponding logical propositions, then we have the correspondence shown in Table 1. In Table 1, $\theta_1 = \overline{\theta_2}$ means that $\theta_1$ is the set-theoretic complement of $\theta_2$.

---
[1] We often represent a clause not as a disjunction of literals (e.g. $\overline{x_1} \vee x_2$) but as an implication ($x_1 \Rightarrow x_2$). This is done to unambiguously identify which side of the implication the literals are on.

Table 1: Correspondence of set theoretic and logic theoretic notions

| SET THEORETIC | LOGIC THEORETIC |
|---|---|
| $\theta_1 \cap \theta_2$ | $x_1 \wedge x_2$ |
| $\theta_1 \cup \theta_2$ | $x_1 \vee x_2$ |
| $\theta_1 \subset \theta_2$ | $x_1 \Rightarrow x_2$ |
| $\theta_1 = \overline{\theta_2}$ | $x_1 = \neg x_2$ |

In this paper we summarize this correspondence,[2] comparing and contrasting the manipulation of DS Belief functions with certain logic-theoretic manipulations.

An important difference between Propositional Logic and DS Theory is the notion of contradiction. Logic has no notion of *contradiction* other than that of a literal and its negation both being assigned $t$. In contrast, DS Theory can encode conflicts between two arbitrary propositions, corresponding to the logical clause $x_i \wedge x_j \Rightarrow \top$, where $\top$ denotes a contradiction.

In addition, classical logic traditionally assumes a fixed set of clauses. DS Theory can be used to pool multiple bodies of evidence, necessitating a labile set of database clauses. We show the changes necessary to update a database consisting of propositional logic clauses.

### 4.1 Symbolic Belief Function Computation

We now show the correspondence of set theoretic notions and propositional clauses, of symbolic Belief functions and minimal support clauses, and of the Belief function update rule $\bigoplus$ and support clause updating.

We start out by defining a set of DS Theory focal propositions $\Theta = \{\theta_1, ..., \theta_n\}$ and corresponding propositional logic propositions (or literals) $\mathbf{x} = \{x_1, ..., x_n\}$. To each focal proposition there is a function $\varrho : \theta \rightarrow [0, 1]$ or $\varrho : x \rightarrow [0, 1]$ which assigns mass to the proposition. We define a set of clauses $X = \{X_1, ..., X_m\}$ which denote the provability relations underlying $\mathcal{P}\Theta$.

First, we define what evaluating the mass assigned to a support clause means:

**Definition:** The mass assigned to a support clause $\xi(x, X)$ is given by

$$\varrho(\xi(x, X)) = \prod_{x_j \in \xi} \varrho(x_j). \quad (3)$$

For example, for a support clause $\xi(x_7, X) = \overline{x_2} \vee \overline{x_4}$, we have $\varrho(\xi(x_7, X)) = \varrho(x_2) \cdot \varrho(x_4)$.

The support clause for a literal is equivalent to a symbolic representation of the Belief assigned to that literal:

**Lemma 1**

$$Bel(\theta) = \sum_{\substack{\theta_i \in \theta \\ \theta_i \neq \emptyset}} \varrho(\theta_i) \iff Bel(x) = \sum_i \varrho(\xi_i(x, X))$$

Given a fixed database, i.e. a fixed set $X$ of clauses, the Belief assigned to any literal or subset of literals can be symbolically computed from the set of support for the literal or subset of literals. DS Theory can also be used in the case of pooling several bodies of evidence, which is equivalent to changing the fixed set of clauses $X$. Belief function updating is necessary in pooling bodies of evidence. In a logical framework, a database can be incrementally updated by support clause updating. For example, if the database is updated by a clause $x_5 \wedge x_7 \Rightarrow x_{new}$ such that $x_5, x_7 \in \mathbf{x}$ and $x_{new} \notin \mathbf{x}$, then the set of support for $x_{new}$ can be incrementally computed from the sets of support for $x_5$ and $x_7$. Thus, if we have $x_5 \wedge x_7 \Rightarrow x_{new}$, and $x_5$ and $x_7$ have support sets $\{\{x_1, x_2\}, \{x_2, x_3\}\}$ and $\{\{x_1\}, \{x_4, x_6\}\}$ respectively, then $x_{new}$ is assigned the support set $\{\{x_1, x_2\}, \{x_2, x_3, x_4, x_6\}\}$ by taking a set union of the support sets for $x_5$ and $x_7$. (See [3] or [11] for a full description of such updating using an ATMS.)

---
[2] Described fully in [12].

We now show the correspondence between Belief function updating and support clause updating. In DS Theory, Belief function updating is done according to Dempster's Rule of Combination (equation 2), and is summarized as $Bel(\theta) = \bigoplus_i Bel(\theta_i)$. Support clause updating must be done to compute the support clause for a newly-introduced literal $x$ if there are support clauses $\xi(x_j, X), ..., \xi(x_k, X)$ such that $\bigwedge_i x_i \Rightarrow x$. The correspondence between DS and logical updating is given by:

**Lemma 2**
$$Bel(\theta) = \bigoplus_i Bel(\theta_i) \iff Bel(x) = \bigwedge_i (\xi_i(x, X)),$$

where $\xi(x_i, X)$ is the support set for $x_i$ with respect to $X$ such that $\bigwedge_i x_i \Rightarrow x$.

Computing Belief for subsets of $\Theta$ is equivalent to computing the set $\Pi(X)$ and from $\Pi(X)$ deriving $\xi(Y, X)$ for $Y$ a literal or clause. This provides only symbolic Boolean expressions for the Belief functions, and these must be evaluated. In general, a Boolean expression is not necessarily disjoint (that is, each pair of disjuncts is disjoint), and a disjoint expression is necessary for the evaluation of the correct Belief assignment. The Boolean expression must be expanded if it is not disjoint, a process which corresponds to what is known in the literature as a Network Reliability computation.

### 4.2 Network Reliability Computation

The Network Reliability problem can be described as follows. The input is a Boolean expression $F$ (which describes a network in which each literal represents a network component) and an assignment of weights to Boolean variables $\varrho : x \to [0, 1]$ (which corresponds to the probability that the component $x$ is functioning). The network reliability problem is to compute the probability that the network (or a portion of the network) is functioning. If we frame this problem in graph theoretic terms, the weighted Boolean expression corresponds to a weighted graph. Hence, the network reliability problem in graphical terms is computing the probability that a set $V$ of vertices can communicate with one another (i.e. that a set of paths exists between the vertex set $V$). The set of support corresponds to the set of paths/cutsets (for $F$ expressed in DNF/CNF respectively) of a graph. Hence network reliability can be computed directly from the graph $\mathcal{G}$ or from the paths or cutsets of $\mathcal{G}$.

This correspondence between computing DS Belief functions and computing network reliability is useful because the latter problem has been carefully studied for many years, and results derived by the network reliability literature can be used for DS theory computations.

Numeric assignments of Belief can be calculated as given by:

**Lemma 3** *The DS Belief assigned to a literal can be computed using an ATMS by converting the weighted ATMS label set to its graphical representation and computing the probability that an $s - t$ path exists in the subgraph formed from the label assigned to the literal.*

Hence calculating DS belief functions for an underlying Boolean expression $F$ is identical to computing the network reliability for the graph corresponding to $F$.

Several methods have been developed for computing network reliability. These methods and their applicability to DS Belief function computation are described in [11].

## 5 ATMS-BASED IMPLEMENTATION OF DEMPSTER SHAFER THEORY

We call an ATMS-based implementation of Dempster Shafer theory an extended ATMS.[3] In describing this implementation, we need to introduce some ATMS terminology. The ATMS is a database management system which, given a set $X$ of propositional clauses, computes a set of support (called a label, $\mathcal{L}(x)$) for each database literal $x$. $\mathcal{L}(x)$ consists only of assumptions, a distinguished subset of the database literals. The assumptions, which we denote by $\mathcal{A} = \{A_1, ..., A_l\}$, are the primitive data representation of the ATMS. The labels for literals thus summarize "proofs" in terms of a Boolean expression consisting of assumptions only. In logical terms, an ATMS label is a restriction of the support set (defined earlier) to assumptions. The ATMS-based implementations assign mass only to assumptions. Additionally they are restricted to Horn clauses, as the ATMS slows considerably with non-Horn clauses.

The ATMS records contradictions in terms of a conjunction of assumptions called a *nogood*. By ensuring null intersections of all labels with the set of nogoods, the ATMS maintains a consistent assignment of labels to database literals. The ATMS can incrementally update the database labeling following the introduction of new clauses. It does this by storing the entire label and nogood set to avoid computing them every time they are needed.

Belief can be assigned only to non-contradictory subsets. In probabilistic terms, this corresponds to conditioning on non-contradictory evidence. Conditioning in DS theory is expressed by Dempster's Rule of Conditioning:

**Lemma 4** *If Bel and Bel' are two combinable Belief*

---
[3]Similar implementations have been done by Laskey and Lehner [6] and d'Ambrosio [2].

*functions*,[4] *let* $Bel(\cdot|\theta_2)$ *denote* $Bel \oplus Bel'$. *Then*

$$Bel(\theta_1|\theta_2) = \frac{Bel(\theta_1 \cup \overline{\theta_2}) - Bel(\overline{\theta_2})}{1 - Bel(\overline{\theta_2})} \qquad (4)$$

for all $\theta_1 \subset \Theta$.

There is an analog in the ATMS to Dempster's rule of Conditioning.

**Lemma 5** *If we call the set of nogoods* $\Psi$, *then the ATMS's symbolic representation of equation 4 is, for all* $x \in \mathbf{x}$,

$$Bel(x \mid \neg\Psi) = \frac{Bel[\mathcal{L}(x) \cup \Psi] - Bel[\Psi]}{1 - Bel[\Psi]}. \qquad (5)$$

It is immediately obvious that the ATMS can be used to compute the symbolic representation of Belief functions as described earlier. We given a brief description of the algorithm, and refer the reader to the relevant papers (Provan [11], Laskey and Lehner [6] and d'Ambrosio [2]).

**ATMS-based Belief Function Algorithm**

1. Compute a Boolean expression from the label: $\mathcal{L} = \bigvee_i L_i$, where each $L_i = \bigwedge_k A_k$ for the set of $k$ assumptions.

2. Account for nogoods, using equation 5.

3. Convert the Boolean expression (5) into a disjoint form (a Network Reliability computation).

4. Substitute mass functions for the $A_i$'s to calculate the mass function for $x$.

Example 1:
Consider a following example with nogoods: the set of clauses is (represented both as implications and Horn clauses)

$x_1$
$x_4$
$x_1 \wedge A_1 \Rightarrow x_2$     $\overline{x_1} \vee \overline{A_1} \vee x_2$
$x_2 \wedge A_2 \Rightarrow x_3$     $\overline{x_2} \vee \overline{A_2} \vee x_3$
$x_1 \wedge A_3 \Rightarrow x_4$     $\overline{x_1} \vee \overline{A_3} \vee x_4$
$x_4 \wedge A_4 \Rightarrow x_5$     $\overline{x_4} \vee \overline{A_4} \vee x_5$
$x_2 \wedge x_4 \wedge A_5 \Rightarrow x_5$     $\overline{x_2} \vee \overline{x_4} \vee \overline{A_5} \vee x_5$

The masses assigned to the assumptions are:

| ASSUMPTION | MASS |
|---|---|
| $A_1$ | .5 |
| $A_2$ | .7 |
| $A_3$ | .8 |
| $A_4$ | .6 |
| $A_5$ | .9 |
| $A_6$ | .4 |

The labels the ATMS assigns to the literals are:

| LITERAL | LABEL |
|---|---|
| $x_2$ | $\{A_1\}$ |
| $x_3$ | $\{A_1, A_2\}$ |
| $x_4$ | $\{A_3\}$ |
| $x_5$ | $\{\{A_1, A_5\}, \{A_1, A_3, A_4\}\}$ |

---

[4]cf. [14], p.67 for a definition of conditions for combinability.

The computation of the Boolean expressions for (and hence Belief assigned to) these labels is trivial except for the expressions for $x_5$, which we now show:

$$\begin{aligned} Bel(x_5) &= \varrho\Big(\{\{A_5, A_1\}, \{A_1, A_3, A_4\}\}\Big) \\ &= \varrho\big((A_5 \wedge A_1) \vee (A_1 \wedge A_3 \wedge A_4)\big) \\ &= \varrho\big(A_1 \wedge (A_5 \vee A_3 \wedge A_4)\big) \\ &= \varrho(A_1)\varrho(A_5 \vee A_3 \wedge A_4) \\ &= \varrho(A_1)\big(\varrho(A_5) + \varrho(A_3)\varrho(A_4) \\ &\quad -\varrho(A_5)\varrho(A_3)\varrho(A_4)\big) \end{aligned}$$

The Belief assigned to the literals is:

| LITERAL | BELIEF |
|---|---|
| $x_2$ | .5 |
| $x_3$ | .35 |
| $x_4$ | .8 |
| $x_5$ | .51 |

Example 2:
Consider the introduction of a new clause $x_2 \wedge A_6 \Rightarrow \overline{x_6}$, such that $\varrho(A_6) = 0.4$. Suppose we are given the information that $x_4$ and $\overline{x_6}$ are contradictory, so that a nogood is formed:

$$\begin{aligned} \Psi &= \mathcal{L}(x_4) \wedge \mathcal{L}(\overline{x_6}) \\ &= \{A_3\} \wedge \{A_1, A_6\} \\ &= \{A_1, A_3, A_6\} \end{aligned}$$

The new assignment of Belief to literals is:

| LITERAL | BELIEF |
|---|---|
| "nogood" | .16 |
| $x_2$ | .4 |
| $x_3$ | .30 |
| $x_4$ | .76 |
| $x_5$ | .43 |

# 6 MODEL-BASED VISUAL RECOGNITION USING AN EXTENDED ATMS

Provan [10] describes a model-based visual recognition system called VICTORS[5] based on an ATMS. VICTORS was designed to test the use of a logical representation for high level vision, and the use of an ATMS to propagate the set $X$ of logical clauses and maintain consistency within $X$. VICTORS exhibits many novel features: it can simultaneously identify all occurrences of a given figure within a scene of randomly overlapping rectangles, subject to variable figure geometry, input data from multiple sources and incomplete figures. Moreover, it conducts sensitivity analyses of figures, updates figures given new input data without having to entirely recompute the new figures, and is robust given noise and occlusion. A sample image which VICTORS interprets is shown in Figure 1. Artificial input data was used, as real input data distracted from the primary objectives of studying the use of logic and of the ATMS in vision.

---

[5]The acronym stands for Visual Constraint Recognition System.

Figure 1: Scene of Overlapping Rectangles with Several Puppet Interpretations

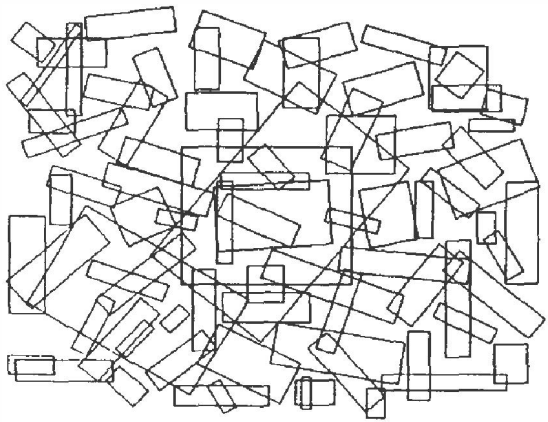

However, the basic implementation of VICTORS suffers from a major deficiency, namely its inability to rank visual interpretations. This is due to the TMS assigning only binary "weights"—each figure part hypothesis is either "believed" or "not believed". Since visual systems typically identify a single best interpretation, this is a major flaw. In addition, the inability to rank interpretations leads to system inefficiency, especially in images with some degree of ambiguity (cf. [9]). This is because ambiguity leads to exploration of a large number of partial interpretations, several of which are definitely not optimal, and should not be explored.

The extension of the ATMS with Belief functions has enabled the weighting of interpretations, thus overcoming this deficiency. We briefly describe this basic implementation, and the assignment of weights in VICTORS, full descriptions of both of which are given in [12].

### 6.1 Basic VICTORS Description

The problem VICTORS solves is as follows: given a set of $n$ 2D randomly overlapping rectangles and a relational and geometric description of a figure, find the best figures if any exist. We define the figure using a set of constraints over the overlap patterns of $k \leq n$ rectangles. The type of figure identified, a puppet consisting of 7 or more parts, is shown in Figure 2. VICTORS can detect any type of object; all it needs is a description of the object encoded as a set of constraints over a set of rectangles. The choice of a puppet as a figure for identification is not central to the operation of VICTORS or the issues it addresses. A puppet is one of many possible figures which fulfills the objectives of (1) being broken up naturally into multiple parts (ranging in the puppet from 7 up), and (2) having interpretations with the subparts taking on various configurations. Such an object model allows great variability in the degree of model complexity specified, and the ability to test the effect of that complexity on the size of search space generated.

VICTORS consists of two main modules, a domain dependent Constraint Engine and a domain independent Reasoning Engine. The Constraint Engine uses a set of constraints for a given figure. A constraint is a set of filters, where each filter is a test of the geometric properties of a set of rectangles. Each constraint places restrictions on acceptable assignments of puppet parts to rectangles based on the overlap patterns of the rectangles. For example, one of the filters for a trunk is that there are at least 5 smaller rectangles overlapping it (which could be a neck and four limbs). We discuss some criteria defining a thigh in § 6.2.1.

Based on the constraint set, the Reasoning Engine generates a set of TMS-clauses, where a TMS-clause is a logical clause which encodes a successful constraint. Each TMS-clause consists of assumptions and TMS-nodes, where a TMS-node is a rectangle/puppet part hypothesis. For example, a TMS-node could be $C : trunk$, and a TMS-clause $A_1 \wedge C : trunk \Rightarrow D : thigh$, where $A_1$ is an assumption. The TMS propagates the set of TMS-clauses to create a set of TMS-nodes. The TMS maintains consistency within this set of TMS-nodes subject to the TMS-clause set. A figure is identified from a consistent set of TMS-nodes which together define the figure.

We present an example to demonstrate the details of the operation of VICTORS in the simplest case of identifying an unambiguous puppet with all parts of the puppet present. As a rule, in figures displaying puppets, most extraneous rectangles are removed so that the points we are stressing in the figures will be clearly evident. In general, scenes are much more cluttered. We refer the reader to [10] and [12] for descriptions of further system capabilities, such as identifying puppets with ambiguous interpretations, missing pieces, occluded pieces, puppets amid clutter, etc.

Example:
Consider the simple task of finding 15-element puppets from a scene of randomly overlapping rectangles, as shown in Figure 2.

Figure 2: Process of Detecting Puppet

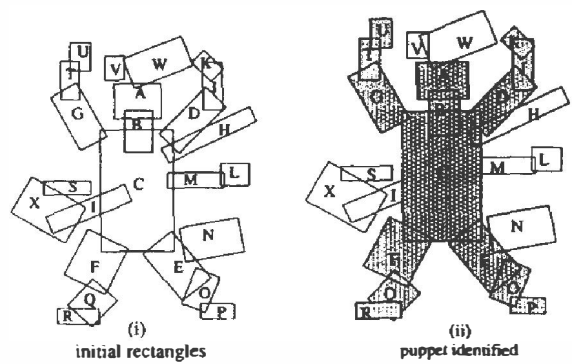

(i) initial rectangles     (ii) puppet identified

First, the Constraint Engine assigns rectangles as *seeds*, where a *seed* is a rectangle/puppet-part hypothesis used to start the growth of puppet figures. In this case the seeds are $A : head$ and $C : trunk$.

Second, starting from the seed rectangles, all subse-quent assignments of puppet part-hypotheses to rectangles are made. Assignments are based on rectangle overlaps and the puppet topology. For example, an overlap with a rectangle identified as a *head* will produce only a *head* − *neck* TMS-clause, and not a *head* − *thigh* TMS-clause, because the head is attached only to the neck. Thus, in Figure 2, seed rectangle $C$ propagates to all its overlaps, which can be several possible combinations of the limbs neck, upper arms and thighs. For example, it propagates *left upper arm* to rectangles $D$ and $H$, since either of these rectangles could eventually end up with that part assignment. $D$:*left-upper-arm* then propagates the part assignment *left-forearm* to rectangle $J$, which in turn propagates the part assignment *left-hand* to rectangle $K$.

Third, the TMS propagates the clauses in the TMS-clause set to produce a set of consistent TMS-nodes. Propagation proceeds as follows: From the TMS-clauses $A$ : *head* (identified as a seed) and $A$ : *head* $\Rightarrow B$ : *neck*, the TMS infers $B$ : *neck*. The TMS continues this propagation process, eliminating multiple and/or contradictory hypotheses (for example $A$ : *head* and $A$ : *neck* are contradictory hypotheses), until a globally consistent set of hypotheses is assigned. The Constraint Engine then takes the rectangle/puppet-part hypothesis set and interprets it as puppet figures. In this case, a full puppet is identified, as shown by shaded rectangles in Figure 2(b).

Note that each (partial) interpretation is associated with an assumption set. If assumptions are now shown in this example, for the partial interpretation derived from the clause set $\{A_1 \Rightarrow A$ : *head*, $A_2 \Rightarrow C$ : *trunk*, $A_3 \wedge C$ : *trunk* $\Rightarrow F$ : *right* − *thigh*, $A_4 \wedge A$ : *head* $\Rightarrow B$ : *neck*$\}$, the assumption set $\{A_1, A_2, A_3, A_4\}$ is obtained for the partial puppet interpretation consisting of $A$ : *head*, $B$ : *neck*, $C$ : *trunk*, $F$ : *right* − *thigh*.

## 6.2 Uncertainty Representation in VICTORS

As mentioned earlier, the basic VICTORS implementation suffers from the inability to rank the interpretations, and outputs a set of interpretations with no way of choosing among them. Extending the ATMS with DS Belief functions enables this ranking to be done, as we now briefly explain.

The ATMS is extended by assigning [0,1] weights to assumptions. In VICTORS, each assumption corresponds to the hypothesis of a rectangle representing a particular seed puppet part, such as $A$ : *head*, or the hypothesis of a TMS-clause, such as $A$ : *head* $\Rightarrow B$ : *neck*. With the assumption explicitly represented we have $A_1 \Rightarrow A$ : *head* and $A_4 \wedge A$ : *head* $\Rightarrow B$ : *neck*. In the process of generating an interpretation for an image, a sequence of assumptions is made, starting from seed assumptions and continuing to the extremities (hands, feet) of the puppet. We now describe the assignment of weights to assumptions.

Weight assignment does not require significantly more processing than is necessary with the traditional ATMS. This is because the rectangle data that exists already is used to define criteria for "quality" of part acceptability. Thus, instead of testing a constraint that the overlap of rectangle $C$, identified as *trunk*, with rectangle $D$ either qualifies $D$ to be a *thigh* or not, a weight or probability with which the constraint could be true is calculated. Hence, we extend the basic VICTORS hypothesis (e.g. $D$ satisfies a constraint to be a *thigh* given rectangle $C$ is hypothesised as a *trunk*) to a weighted hypothesis.

We have been studying the effectiveness of the simplest weight assignments, using more complicated assignments only when necessary. In the following section we present a weight assignment method which approximates more theoretically correct methods and which has been successful for simple input data.

### 6.2.1 Weight assignments

Figure-part hypotheses (e.g. rectangle $D$ being a thigh) are based on rectangle overlaps (e.g. the overlap of $D$ with a rectangle $C$ already assumed to be a *trunk*). Some of the filters which define the constraints governing hypotheses include: (1) angle of overlap; (2) relative area; (3) relative overlap area; and (4) axial ratio. Each filter is satisfied with a [0,1] degree of acceptability; 0 is unacceptable and 1 is perfectly acceptable. In general, there is a probability distribution $\varphi$ over the filter's feasible range. The simplest approximation to $\varphi$ is to define a subset of each filter's range with which the filter is satisfied with high probability, and the remaining subset with low probability. For example, for the thigh, we have the following ranges:

**angle of overlap** As shown in Figure 3(a), the total angular range within which an acceptable overlap occurs is $[\pi, \pi/4]$. We define a sub-range, namely $[5\pi/4, 0]$, as an overlap acceptable with high probability, and the remaining sub-range, $[0, \pi/4]$ and $[\pi, 5\pi/4]$, as an overlap acceptable with low probability. These regions are shown in Figure 3(b). The angle of overlap $\alpha$ is computed to determine acceptability or unacceptability in basic VICTORS. In this extended system, all that is necessary in addition is to place this angle $\alpha$ in the high or low probability category.

**relative area** For acceptability of the trunk-thigh overlap, the ratio of the area of the thigh to the area of the trunk must fall within the bounds $[0.6, 0.15]$. The bounds $[0.4, 0.25]$ define an overlap acceptable with high probability, and the bounds $[0.6, 0.4]$, $[0.25, 0.15]$ define an overlap acceptable with low probability.

**relative overlap area** Similar to the relative area filter, there is a low and high probability ratio of overlap areas. For the thigh and trunk rectangles, this is given by Table 2.

Similar high and low probability assignments exist for the axial ratio and all other filters.

Next, the weights for all the separate criteria must be merged to give an overall weight. Because the criteria correspond to different frames of reference, *refinement* (cf. [14]) is necessary to map these disparate frames onto a common frame, so the the weights from each criterion can be combined. A rough approximation to this refinement process can be obtained as follows. A probability $p_1$ is assigned to the high probability value, and $p_2$ is assigned to the low-probability value. The $p_i$'s for a given

Figure 3: Regions with probabilistic weights of acceptability

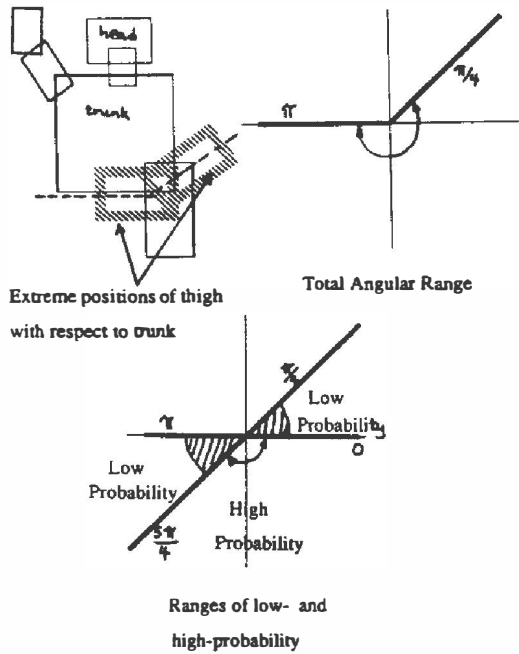

Extreme positions of thigh with respect to trunk

Total Angular Range

Ranges of low- and high-probability

Table 2: Probability assignments to ratio of overlap areas

| THIGH FILTER TYPE | RATIO |
|---|---|
| Simple acceptability | [0.5, 0] |
| high probability acceptability | [0.3, 0.1] |
| low probability acceptability | [0.5, 0.3], [0.1, 0] |

| TRUNK FILTER TYPE | RATIO |
|---|---|
| Simple acceptability | [0.25, 0] |
| high probability acceptability | [0.15, 0.05] |
| low probability acceptability | [0.25, 0.15], [0.05, 0] |

assumption are multiplied together and normalized to ensure that the highest acceptability weight assigned is 1. Thus, if four filters are used to define the constraint for thigh acceptability, and $p_1 = 0.8$ and $p_2 = 0.5$, the normalization constant is $p_1^4 = 0.8^4 = 0.4096$. If we have 3 high-probability values and 1 low-probability values, the weight assigned is given by $(0.8^3 \times 0.5)/0.4096$, which works out to 0.625. Some weights obtained based on different combinations of high- and low-probability criteria are given in Table 3.

The methods of assigning weights, and the values of weights themselves are somewhat arbitrary. What is needed is a *theory* of assigning weights, and of learning appropriate assignments. Lowe [8], for example, discusses some criteria necessary for such a theory, and Binford et al. [1] propose a theory based on quasi-invariants. However, much more work needs to be done.

### 6.3 Results

Given the assignment of weights to assumptions, the DS Belief functions of interpretations are computed as described in previous sections.

The use of DS Belief functions has enabled a ranking of interpretations, meaning that the best interpretation can be found. We show how this comes about with an example. Figure 4(a) shows an input image. Figure 4(b) shows some interpretations which can be discovered using VICTORS with a traditional ATMS. Figure 4(c) shows the best interpretation found by VICTORS with an extended ATMS.

Additionally, we are studying different methods of using this ranking to prune the search space by exploring only the best partial interpretations. This has the potential of enhancing the efficiency of VICTORS.

Results to date indicate that even simple weight assignments prove useful in generating an ordering of partial interpretations equivalent to the theoretically accurate ordering. However, for more complicated input data these simple techniques are too inaccurate. Indeed, we anticipate that real, sensor-derived data will require sophisticated weight manipulation. Even so, there are domains in which simple weight assignments can provide the partial ordering necessary for directing search and improving the efficiency of the ATMS. Where appropriate, these computationally efficient approximations can replace the more computationally intensive DS representations.

### 6.4 Related Work

VICTORS is related to the system of Hutchinson et al. [5] in that both systems use DS theory for model-based object recognition. Major differences include the use of 3D range data by [5] in contrast to the synthetic data of VICTORS, and the use of DS theory to enforce relational constraints in [5] as opposed the use of logic in VICTORS. VICTORS is also related to the system of Binford et al. ([1], [7]) in its use of an uncertainty calculus for model-based object recognition, except that [7] uses a probability-based influence diagram representation.

## 7 DISCUSSION

The relation between DS Theory and propositional logic has been described. We have shown how the support clause $\xi(X_i, X)$ gives a notion of a symbolic explanation for $X_i$. In the same way, a symbolic representation for a DS Belief function provides a notion of a symbolic explanation. Moreover, the numeric value of the Belief can be viewed as a numeric summary (or as the believability) of that explanation. In addition, just as a logical model describes which propositions are true in a given world, the DS Belief assigned to propositions describes the degree to which that set of propositions is true. Thus, to the extent to which logic and DS Theory overlap, DS

Table 3: Weight assignments to thigh assumption

| PROBABILITY TYPES | WEIGHT |
|---|---|
| 4 high | 1.0 |
| 3 high, 1 low | 0.625 |
| 2 high, 2 low | 0.4 |
| 4 low | 0.15 |

Theory can acquire a logical semantics. Note that DS Theory has a different notion of contradiction to logic, in that two arbitrary propositions can be defined (external to the logic) as being contradictory.

We have described an application of an ATMS extended with DS Belief functions to visual interpretation. For domains in which the best interpretation is required and truth maintenance is important, such an approach appears promising.

## ACKNOWLEDGEMENTS

I would like to thank Judea Pearl for helpful discussions, and Alex Kean for much help with the Logic-based DS Theory implementation.

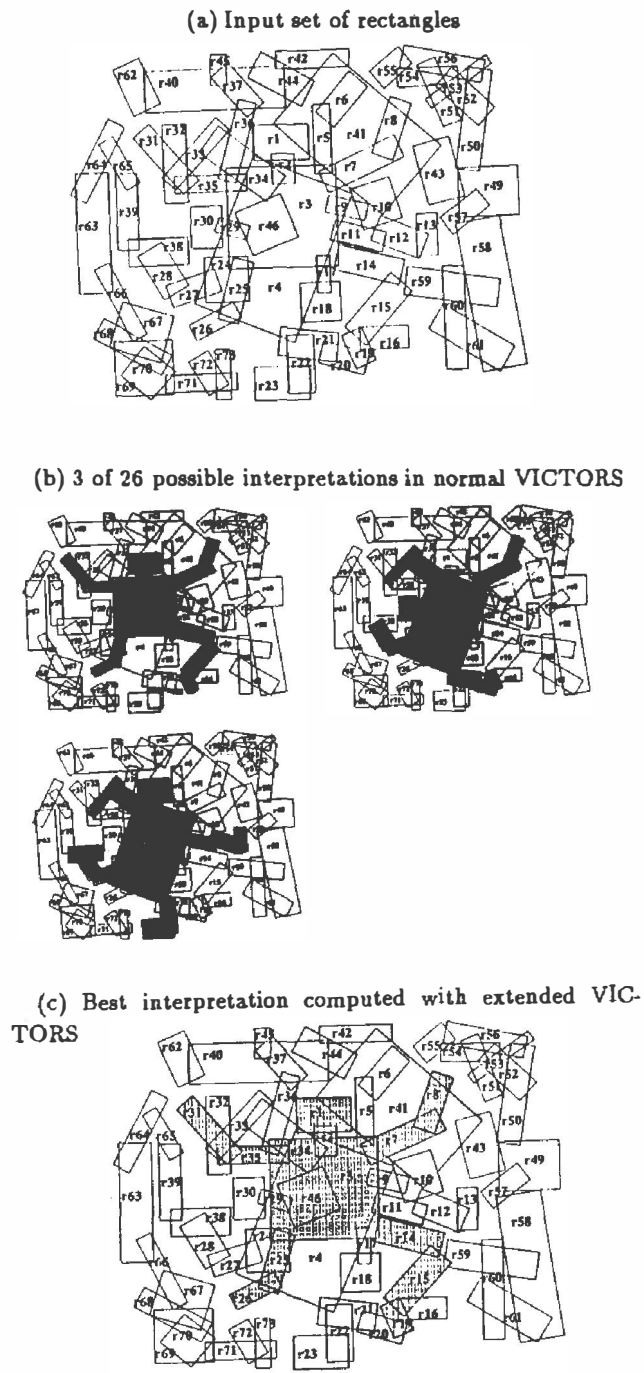

Figure 4: Interpretations found by VICTORS with traditional and extended ATMS

(a) Input set of rectangles

(b) 3 of 26 possible interpretations in normal VICTORS

(c) Best interpretation computed with extended VICTORS

294